\documentclass[11pt,a4paper]{article}
\usepackage[hyperref]{acl2017}
\usepackage{times}
\usepackage{latexsym}

\usepackage{url}

\usepackage{graphicx}
\usepackage{amssymb,amsmath,bm}
\usepackage{textcomp}
\usepackage{multirow}
\usepackage{dsfont}
\usepackage{color}
\usepackage[normalem]{ulem}
\usepackage{rotating}
\usepackage{booktabs}
\usepackage{makecell}
\usepackage{subcaption}
\usepackage{comment}
\usepackage{amsmath,amsfonts,amssymb}
\usepackage{tikz}
\usepackage{verbatim}
\usepackage{caption}
\usepackage{booktabs}
\usepackage{ctable}
\usepackage{makecell}
\usepackage{array} 
\usepackage{varwidth} 
\usepackage{stmaryrd}
\usepackage{enumerate}

\newcommand*{\Scale}[2][4]{\scalebox{#1}{$#2$}}%
\hypersetup{draft}

\aclfinalcopy 
\def\aclpaperid{***} 

\title{Sample-efficient Actor-Critic Reinforcement Learning \\with Supervised Data for Dialogue Management}

\author{
Pei-Hao Su, Pawe\l{} Budzianowski,  Stefan~Ultes, Milica Ga{\v s}i{\' c}, and Steve Young \\
Department of Engineering,
University of Cambridge,
Cambridge, UK\\
\texttt{\{phs26, pfb30, su259, mg436, sjy11\}@cam.ac.uk} \\
}

\date{}

\begin{document}
\maketitle

\begin{abstract}
Deep reinforcement learning (RL) methods have significant potential for dialogue policy optimisation. 
However, they 
suffer from a poor performance in the early stages of learning. This is especially problematic for on-line learning with real users.
Two approaches are introduced to tackle this problem. Firstly, to speed up the learning process, two sample-efficient neural networks algorithms: trust region actor-critic with experience replay (TRACER)  and episodic natural actor-critic with experience replay (eNACER) are presented. 
For TRACER, the trust region helps to control the learning step size and avoid catastrophic model changes. 
For eNACER, the natural gradient identifies the steepest ascent direction in policy space to speed up the convergence.
Both models employ  off-policy learning with experience replay to improve sample-efficiency. 
Secondly, to mitigate the cold start issue, a corpus of demonstration data is utilised to pre-train the models prior to 
on-line reinforcement learning. 
Combining these two approaches, we demonstrate a practical approach to learn deep RL-based dialogue policies and
demonstrate their effectiveness in a task-oriented information seeking domain.

\end{abstract}

\section{Introduction} \label{sec:intro}
Task-oriented Spoken Dialogue Systems (SDS) aim to assist users to achieve specific goals via speech, such as hotel booking, restaurant information and  accessing bus-schedules.
These systems are typically designed according to a structured {\it ontology} (or a database {\it schema}), which defines the domain that the system can talk about.
The development of a robust SDS traditionally requires a substantial amount of hand-crafted rules combined with various statistical components. This includes a spoken language understanding module \cite{chen2016end,yang2017end}, a dialogue belief state tracker \cite{Henderson2014b,perez2016dialog,Mrksic:17} to predict user intent and track the dialogue history, a dialogue policy \cite{POMDP-review,GPRL,pfb_hrl} to determine the dialogue flow, and a natural language generator \cite{rieser2009natural,wensclstm15,hu2017controllable} to convert conceptual representations into system responses.

In a task-oriented SDS, teaching a system how to respond appropriately in all situations is non-trivial. 
Traditionally, this {\it dialogue management} component has been designed manually using flow charts.
More recently, it has been formulated as a planning problem and solved using reinforcement learning (RL) to optimise a dialogue policy through interaction with users \cite{levin1997stochastic,roy2000spoken,POMDP_williams,jurvcivcek2011natural}. In this framework, the system learns by a {\it trial and error} process governed by a potentially delayed learning objective called the {\it reward}.  This reward is designed to encapsulate the desired behavioural features of the dialogue. Typically it provides a positive reward for success plus a per turn penalty to encourage short dialogues
\cite{el2014task,Su_2015,vandyke15a,su:2016:acl}.

To allow the system to be trained on-line, Bayesian sample-efficient learning algorithms have been proposed \cite{GPRL,KTD} which can learn policies from a minimal number of dialogues.
However, even with such methods, the initial performance is still relatively poor, and this can impact negatively on the user experience.

Supervised learning (SL) can also be used for dialogue action selection.  In this case, 
the policy is trained to produce an appropriate response for any given dialogue state. Wizard-of-Oz (WoZ) methods \cite{Kelley84,dahlback1993wizard} have been widely used for collecting domain-specific training corpora.
Recently an emerging line of research has focused on training neural network-based dialogue models, mostly in text-based systems \cite{vinyals2015neural,shang2015neural,serban2015hierarchical,wenN2N16,fb_n2n}.
These systems are directly trained on past dialogues without detailed specification of the internal dialogue state.
However, there are two key limitations of using SL in SDS. Firstly, the effect of selecting an action on the future course of the dialogue is not considered and this may result in sub-optimal behaviour. Secondly, there will often be a large number of dialogue states  which are not covered by the training data \cite{henderson2008hybrid,li2014temporal}. Moreover, there is no reason to suppose that the recorded dialogue participants are acting optimally, especially in high noise levels.
These problems are exacerbated in larger domains where multi-step planning is needed. 

In this paper, we propose a network-based approach to policy learning which combines the best of both SL- and RL-based dialogue management, and which capitalises on recent advances in deep RL \cite{mnih2015human}, especially off-policy algorithms
\cite{wang2016sample}.

The main contribution of this paper is two-fold:
\begin{enumerate}
\item improving the sample-efficiency of 
actor-critic RL:
trust region actor-critic with experience replay (TRACER) and episodic natural actor-critic with experience replay (eNACER).
\item efficient utilisation of demonstration data for improved early stage policy learning.
\end{enumerate}

The first part focusses primarily on increasing the RL learning speed. 
For TRACER, trust regions are introduced to standard actor-critic to control the step size and thereby avoid catastrophic model changes. For eNACER, the natural gradient identifies steepest ascent direction in policy space to ensure fast convergence. Both models exploit the off-policy learning with experience replay (ER) to improve sample-efficiency. These are compared with various state-of-the-art RL methods.

The second part aims to mitigate the cold start issue by using  
{\it demonstration data} to pre-train an RL model.
This resembles the training procedure adopted in recent game playing applications \cite{silver2016mastering,SLRL_atari}. A key feature of this framework is that a single model is trained using both SL and RL with different training objectives but without modifying the architecture.

By combining the above, we demonstrate a practical approach to learning deep RL-based dialogue policies for new domains which can achieve competitive performance without significant detrimental impact on users.

\section{Related Work} \label{sec:relate} 
RL-based approaches to dialogue management have been actively studied for some time \cite{levin1998using,lemon2006evaluating,GPRL}. Initially, systems suffered from slow training, but recent advances in data efficient methods such as Gaussian Processes (GP) have enabled systems to be trained from scratch in on-line interaction with real users \cite{milica_real_users}.
GP provides an  estimate of the uncertainty in the underlying function and a built-in noise model. This helps to achieve highly sample-efficient exploration and robustness to recognition/understanding errors. 

However, since the computation in GP scales
with the number of points memorised, sparse approximation methods such as the {\it kernel span} algorithm \cite{engel2005algorithms} must be used and this limits the ability to scale to very large training sets.   It is therefore questionable as to whether GP can scale to support commercial wide-domain SDS. Nevertheless, GP provides a good benchmark and hence it is included in the evaluation below.

In addition to increasing the sample-efficiency of the learning algorithms, the use of reward shaping has also been investigated in \cite{el2014task,svgm15} to enrich the reward function in order to speed up dialogue policy learning. 

Combining SL with RL for dialogue modelling is not new. \newcite{henderson2008hybrid} proposed a hybrid SL/RL model that, in order to ensure tractability in policy optimisation, performed exploration only on the states in a dialogue corpus. The policy was then defined manually on parts of the space which were not found in the corpus. 
A method of initialising RL models using logistic regression was also described \cite{rieser2006using}.
For GPRL in dialogue, rather than using a linear kernel that imposes heuristic data pair correlation, a pre-optimised Gaussian kernel learned using SL from a dialogue corpus has been proposed \cite{chen2015hyper}. The resulting kernel was 
more accurate on data correlation and 
achieved better performance, however, the SL corpus did not help to initialise a better policy. Better initialisation of GPRL has been studied in the context of domain adaptation by specifying a GP prior or re-using an existing model which is then pre-trained for the new domain \cite{gbhk13}.

A number of authors have proposed training a standard neural-network policy in two stages \cite{fatemi2016policy,su2016continuously,williams2017end}.
\newcite{asadi2016sample} also explored off-policy RL methods for dialogue policy learning.
All these studies were conducted in simulation, using error-free text-based input. A similar approach was also used in a conversational model \cite{li2016deep}.
In contrast, our work introduces two new sample-efficient actor-critic methods, combines both two-stage policy learning and off-policy RL, and testing at differing noise levels.

\section{Neural Dialogue Management } \label{secmodels}
The proposed framework addresses the dialogue management component in a modular SDS.
The input to the model is the belief state $\mathbf{b}$ that encodes a distribution over the possible user intents along with the dialogue history. The model's role is to select the system action $a$ at every turn that will lead to the maximum possible cumulative reward and a successful dialogue outcome.
The system action is mapped into a system reply at the semantic level, and this is subsequently passed to the natural language generator for output to the user.

The semantic reply consists of three parts: the {\it intent} of the response, (e.g. inform), which {\it slots} to talk about (e.g. area), and a {\it value} for each slot (e.g. east). 
To ensure tractability, the policy selects $a$ from a restricted action set which identifies the {\it intent} and sometimes a {\it slot}, any remaining information required to complete the reply is extracted using heuristics from the tracked belief state.

\subsection{Training with Reinforcement Learning}
\label{sec:rl}

Dialogue policy optimisation can be seen as the task of learning to select the sequence of responses (actions) at each turn
which maximises the long-term objective defined by the reward function. 
This can be solved by applying either value-based or policy-based methods.
In both cases, the goal is to find an optimal policy $\pi^*$ that maximises the discounted total return $R = \sum_{t=0}^{T-1} \gamma^t r_t(\mathbf{b}_t, a_t)$ over a dialogue with $T$ turns where $r_t(\mathbf{b}_t, a_t)$ is the reward when taking action $a_t$ in dialogue belief state $\mathbf{b}_t$ at turn $t$ and $\gamma$ is the discount factor. 

The main difference between the two categories is that policy-based methods have stronger convergence characteristics than value-based methods. The latter 
often diverge when using function approximation since they optimise in value space and a slight change in value estimate can lead to a large change in policy space \cite{sutton1999policy}. 

Policy-based methods suffer from low sample-efficiency, high variance and often converge to local optima since they typically learn via Monte Carlo estimation \cite{williams1992simple,schulman2015high}. However, they are preferred due to their superior convergence properties.
Hence in this paper we focus on policy-based methods but also
include a value-based method as a baseline.

\begin{figure}[t]
\centerline{\includegraphics[scale=0.24]{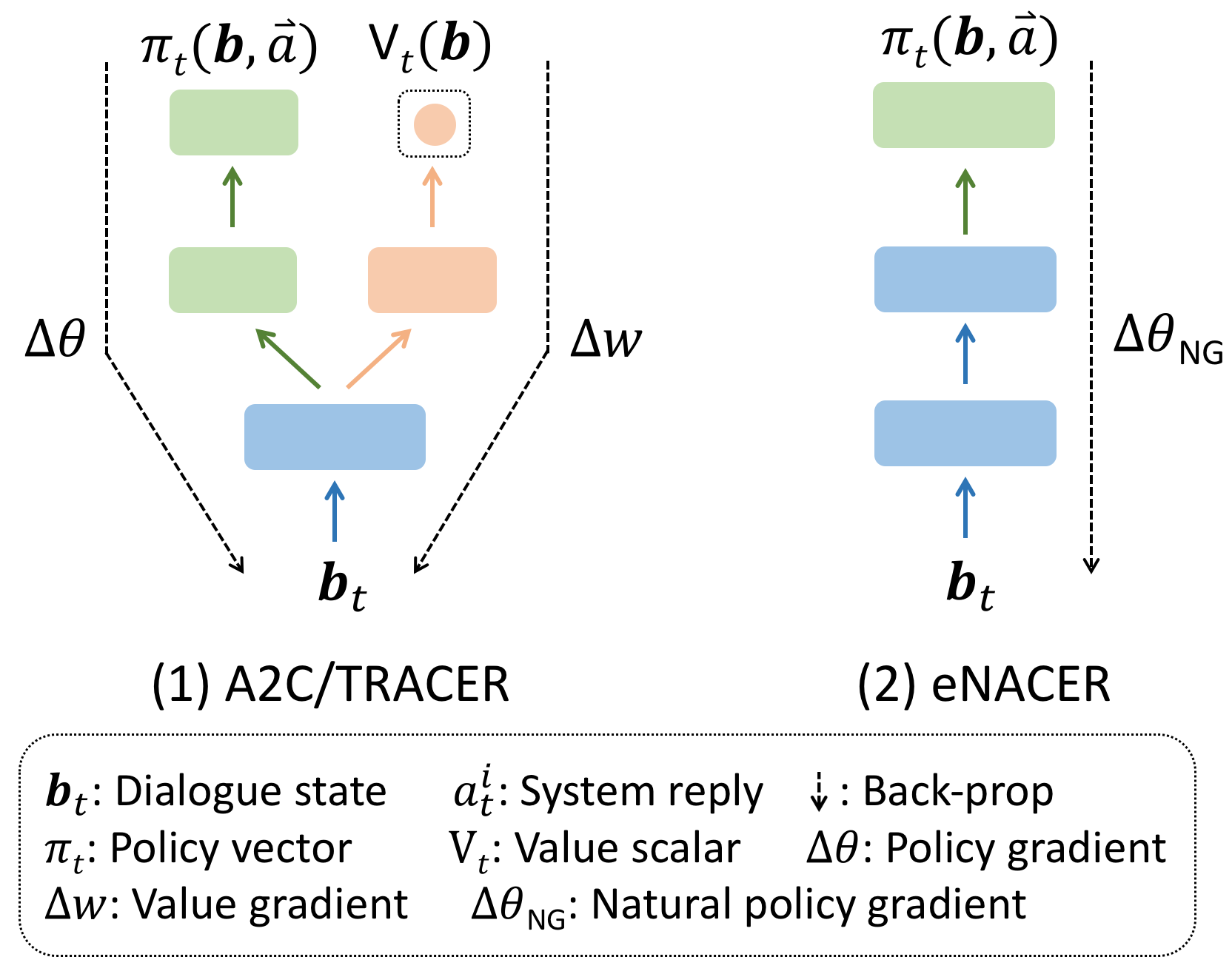}}
\caption{A2C, TRACER and eNACER architectures using feed-forward neural networks.} 
\label{fig:DRLs}
\vspace{-4mm}
\end{figure}

\subsubsection{Advantage Actor-Critic (A2C)}
\label{sec:a2c}
In a policy-based method, the 
training objective is to find a parametrised policy $\pi_\theta(a|\mathbf{b})$ that maximises the expected 
reward $J(\theta)$ over all possible dialogue trajectories given a starting state.

Following the {\it Policy Gradient Theorem} \cite{sutton1999policy}, the gradient of the parameters given the objective function has the  form: 
\begin{equation} \label{eq:advanGrad}
\nabla_\theta J(\theta) = \mathbb{E} \left[ \nabla_\theta \log \pi_{\theta}(a|\mathbf{b})Q^{\pi_{\theta}}(\mathbf{b},a) \right].
\end{equation}
Since this form of gradient has a potentially high variance,
a baseline function is typically introduced to reduce the variance whilst not changing the estimated gradient \cite{williams1992simple,RL}. A natural candidate for this baseline is the value function $V(\mathbf{b})$. Equation \ref{eq:advanGrad} then becomes:
\begin{equation} \label{eq:advanGrad}
\nabla_\theta J(\theta) = \mathbb{E} \left[ \nabla_\theta \log \pi_{\theta}(a|\mathbf{b})A_w(\mathbf{b},a) \right],
\end{equation}
where $A_w(\mathbf{b},a) = Q(\mathbf{b},a) - V(\mathbf{b})$  is the {\it advantage function}. 
This can be viewed as a special case of the {\it actor-critic}, where $\pi_{\theta}$ is the actor and $A_w(\mathbf{b},a)$ is the critic, defined by two parameter sets $\theta$ and $w$. 
To reduce the number of required parameters, temporal difference (TD) errors $\delta_w = r_{t} + \gamma V_w(\mathbf{b_{t+1}}) - V_w(\mathbf{b_t})$ can be used to approximate the advantage function
\cite{schulman2015high}. The left part in Figure \ref{fig:DRLs} shows the architecture and parameters of the resulting A2C policy.

\subsubsection{The TRACER Algorithm}
\label{sec:stabilising}

To boost the performance of A2C policy learning, two methods are introduced:
\begin{enumerate}[I.]
\item {\bf Experience replay with off-policy learning for speed-up}
\end{enumerate}

{\it On-policy} RL methods update the model with the samples collected via the current policy. 
Sample-efficiency can be improved by utilising {\it experience replay} (ER) \cite{lin1992self}, where mini-batches of dialogue experiences are randomly sampled from a replay pool $\mathcal{P}$ to train the model.
This increases learning efficiency by re-using past samples in multiple updates whilst ensuring stability by reducing the data correlation.
Since these past experiences were collected from different policies compared to the current policy, the use of ER leads to {\it off-policy} updates.

When training models with RL, $\epsilon$-greedy action selection is often used to trade-off between exploration and exploitation,
whereby a random action is chosen with probability $\epsilon$ 
otherwise the top-ranking action is selected. 
A policy used to generate a training dialogues (episodes) is referred to as a {\it behaviour policy} $\mu$, in contrast to the policy to be optimised which is called the {\it target policy} $\pi$.

The basic A2C training algorithm described in \S \ref{sec:a2c}
is {\it on-policy} since it is  assumed that actions are drawn from the same policy as the target to be optimised 
($\mu = \pi$). 
In {\it off-policy} learning, since the current policy $\pi$ is updated with the samples generated from old behaviour policies $\mu$, an importance sampling (IS) ratio is used to rescale each sampled reward to correct for the sampling bias at time-step $t$: $\rho_t = {\pi(a_t|\mathbf{b_t})}/{\mu(a_t|\mathbf{b_t})}$ \cite{meuleau2000off}.

For A2C, the off-policy gradient for the parametrised value function $V_w$ thus has the form:
\begin{equation}
\Scale[0.90]{\Delta w^{\textrm{off}} = \sum_{t=0}^{T-1} \big( \bar{R}_t-\hat V_w(\mathbf{b_t})\big)\nabla_w\hat V_w(\mathbf{b_t}) \underset{i=0}{\overset{t}{\Pi}}\rho_i,}
\label{eq:off_policy_value}
\end{equation}
where $\bar{R}_t$ is the off-policy Monte-Carlo return \cite{precup2001off}:
\begin{equation}
\Scale[0.8]{ \bar{R}_t =  r_{t} + \gamma r_{t+1} \underset{i=1}{\overset{1}{\Pi}}\rho_{t+i} + \cdots + \gamma^{T-t-1} r_{T-1} \underset{i=1}{\overset{T-1}{\Pi}}\rho_{t+i} }.
\label{eq:off_policy_mc_return}
\end{equation}
Likewise, the updated gradient for policy $\pi_\theta$ is:
\begin{equation}
\label{eq:pg}
\Delta \theta^{\textrm{off}} = \sum_{t=0}^{T-1}\rho_t \nabla_{\bm{\theta}} \log \pi_{\theta}(a_t|\mathbf{b}_t) \hat \delta_w,
\end{equation}
where $\hat \delta_w = r_{t} + \gamma \hat V_w(\mathbf{b_{t+1}}) - \hat V_w(\mathbf{b_{t}})$ is the TD error using the estimated value of $\hat V_w$.

Also, as the gradient correlates strongly with the sampled reward, reward $r_t$ and total return $R$ are normalised to lie in [-1,1] to stabilise training.

\begin{enumerate}[I.]
  \setcounter{enumi}{1}
  \item {\bf Trust region constraint for stabilisation}
\end{enumerate}

To ensure  stability in RL, each per-step policy change is often limited by setting a small learning rate. However, 
setting the rate low enough to avoid occasional large destabilising updates is not conducive to fast learning.

Here, we adopt a modified Trust Region Policy Optimisation method introduced by \newcite{wang2016sample}. In addition to maximising the cumulative reward $J(\theta)$, the optimisation is also subject to a Kullback-Leibler (KL) divergence limit between the updated policy $\theta$ and an {\it average policy}  $\theta_a$ to ensure safety. This average policy represents a running average of past policies and constrains the updated policy to not deviate far from the average $\theta_a \leftarrow \alpha \theta_a + (1- \alpha)\theta$ with a weight $\alpha$.

Thus, given the off-policy policy gradient $\Delta \theta^{\textrm{off}}$ in Equation \ref{eq:pg}, the modified policy gradient with trust region $g$ is calculated as follows: 
\begin{equation*}
\label{eqn:trust}
\begin{aligned}
  & \underset{g}{\text{minimize}}
  & & \frac{1}{2}\| {\Delta \theta^{\textrm{off}}} - g\|^2_2, \\
  & \text{subject to}
  & & \nabla_{{\theta}} D_{KL}\left[\pi_{\theta_a}(\mathbf{b_t}) \| \pi_{\theta}(\mathbf{b_t}) \right]^T g \leq \xi,
\end{aligned}
\end{equation*}
where $\pi$ is the policy parametrised by $\theta$ or $\theta_a$, and $\xi$ controls the magnitude of the KL constraint.
Since the constraint is linear, a closed form solution to this quadratic programming problem can be derived using the KKT conditions. Setting $k = \nabla_{{\theta}} D_{KL}\left[\pi_{\theta_a}(\mathbf{b_t}) \| \pi_{\theta}(\mathbf{b_t}) \right]$, we get:
\begin{equation}
g^*_{tr} = \Delta \theta^{\textrm{off}} - \max\left\{ \frac{ k^T \Delta \theta^{\textrm{off}} - \delta}{\| k\|^2_2},0\right\} k.
\label{eq:new_pg}
\end{equation}
When this constraint is satisfied, there is no change to the gradient with respect to $\theta$. Otherwise, the update is scaled down along the direction of $k$ and the policy change rate is lowered. This direction is also shown to be closely related to the {\it natural gradient} \cite{amari1998natural,schulman2015trust}, which is presented in the next section.

The above enhancements speed up and stabilise A2C. We call it the Trust Region Actor-Critic with Experience Replay (TRACER) algorithm.

\subsubsection{The eNACER Algorithm}
\label{sec:enac}

Vanilla gradient descent algorithms are not guaranteed to update the model parameters in the steepest direction due to re-parametrisation \cite{amari1998natural,martens2014new}.
A widely used solution to this problem is to use a {\it compatible function approximation} for the advantage function
in Equation \ref{eq:advanGrad}: $\nabla_w A_w(b,a) = \nabla_\theta \log \pi_{\theta}(a|b)$, where the update of $w$ is then in the same update direction as $\theta$ \cite{sutton1999policy}. Equation \ref{eq:advanGrad} can then be rewritten as: 
\begin{equation*} \label{eq:naturalGrad}
\begin{aligned}
\Scale[1]{\nabla_\theta J(\theta)} &=\Scale[0.93]{ \mathbb{E} \left[ \nabla_\theta \log \pi_{\theta}(a|\mathbf{b}) \nabla_\theta \log \pi_{\theta}(a|\mathbf{b})^T w \right]} \\
&{= F(\theta) \cdot w},
\end{aligned}
\end{equation*}
where $F(\theta)$ is the Fisher information matrix. This implies $\Delta {\theta_{NG}} = w = F(\theta)^{-1} {\nabla_\theta J(\theta)}$ and it is called the {\it natural gradient}. 
The Fisher Matrix can be viewed as a correction term which makes the natural gradient independent of the parametrisation of the policy and corresponds to steepest ascent towards the objective \cite{martens2014new}.
Empirically, the natural gradient has been found to significantly speed up convergence.

Based on these ideas, the Natural Actor-Critic (NAC) algorithm was developed by \newcite{peters2006policy}. In its episodic version (eNAC), the Fisher matrix does not need to be explicitly computed.
Instead, the gradient is estimated by a least squares method given
the $n$-th episode consisting of a set of transition tuples $\{(\mathbf{b}_t^n, a_t^n, r_t^n)\}_{t=0}^{T_n -1}$:
\begin{equation}
\label{eq:enac}
\Scale[0.85]{ R^n = \left[ \sum_{t=0}^{T_n -1} \nabla_\theta \log \pi_\theta(a_t^i|\mathbf{b}_t^i; \theta)^T \right] \cdot \Delta {\theta_{NG}} + C},
\end{equation}
which can be solved analytically. $C$ is a constant which is an estimate of the baseline 
$V(\mathbf{b})$.

As in TRACER, eNAC can be enhanced by using ER and off-policy learning, thus called eNACER, whereby 
$R^n$ in Equation \ref{eq:enac} is replaced by the off-policy Monte-Carlo return $\bar{R}^n_0$ at time-step $t=0$ as in Equation \ref{eq:off_policy_mc_return}.
For very large models,
the inversion of the Fisher matrix can become prohibitively expensive to compute. Instead, a truncated variant can be used to calculate the natural gradient \cite{schulman2015trust}.

eNACER is structured as a feed forward network with the output $\pi$ as in the right of Figure \ref{fig:DRLs}, updated with natural gradient $\Delta {\theta_{NG}}$. Note that by using the compatible function approximation, the value function does not need to be explicitly calculated. This makes eNACER in practice a policy-gradient method.

\subsection{Learning from Demonstration Data} \label{sec:SLRL}

From the user's perspective, performing RL from scratch will invariably result in unacceptable performance in the
early learning stages. 
This problem can be mitigated by 
an off-line corpus of {\it demonstration data} to bootstrap a policy. 
This data may come from a WoZ collection or from interactions between users and an existing policy.
It can be used in three ways:
A: Pre-train the model, B: Initialise a supervised replay buffer $\mathcal{P}_{sup}$, and C: a combination of the two.

{\bf (A)} For model pre-training, the objective is to `mimic' the response behaviour from the corpus. This phase is essentially standard SL. 
The input to the model is the dialogue belief state $\mathbf{b}$, and the training objective for each sample is to minimise a joint cross-entropy loss $\mathcal{L}(\theta) = -\sum\nolimits_{k} y_k \log(p_k)$ between action labels $y$ and model predictions $p$, where the policy is parametrised by a set $\theta$.

A policy trained by SL on a fixed dataset may not generalise well. In spoken dialogues, the noise levels may vary across conditions and thus can significantly affect performance. Moreover, a policy trained using SL does not perform any long-term planning on the conversation. Nonetheless, supervised pre-training offers a good model starting point
which can then be fine-tuned using RL.

{\bf (B)} For supervised replay initialisation, the demonstration data is stored in a replay pool $\mathcal{P}_{sup}$
which is kept separate from the ER pool used for RL and is never over-written. At each RL update iteration, a small portion of the demonstration data $\mathcal{P'}_{sup}$ is sampled,
and the supervised cross-entropy loss $\mathcal{L}(\theta)$ computed on this data is added to the RL objective $J(\theta)$. Also, an L2 regularisation loss $ \| \cdot \|^2_2$ is applied to $\theta$
to help prevent it from over-fitting on the sampled demonstration dataset. The total loss to be minimised is thus:
\begin{equation}
\mathcal{L}_{all}(\theta) = -J(\theta) + \lambda_1 \mathcal{L}(\theta; \mathcal{P'}_{sup}) + \lambda_2 \| \theta\|^2_2,
\label{eq:loss_all}
\end{equation}
where $\lambda$'s are weights.
In this way, the RL policy is guided by the sampled demonstration data while learning to optimise the total return.

{\bf (C)} The learned parameters of the pre-trained model in method A above might distribute differently from the optimal RL policy and this may cause some performance drop in early stages while learning an RL policy from this model. This can be alleviated by using the composite loss proposed in method B.  A comparison between the three options is included in the experimental evaluation.

\section{Experimental Results}
Our experiments utilised the software tool-kit PyDial \cite{cupydial}, which provides a platform for modular SDS.
The target application is a live telephone-based SDS providing restaurant information for the Cambridge (UK) area. The task is to learn a policy which manages the dialogue flow and delivers requested information to the user.
The domain consists of approximately 100 venues, each with 6 slots out of which 3 can be used by the system to constrain the search (food-type, area and price-range) and 3 are system-informable properties (phone-number, address and postcode) available once a database entity has been found.

The input for all models was the full dialogue belief state $\mathbf{b}$ of size 268 which includes the last system act and distributions over the user intention and the three requestable slots. 
The output includes 14 restricted dialogue actions determining the system intent at the semantic level.
Combining the dialogue belief states and heuristic rules, it is then mapped into a spoken response using a natural language generator.

\subsection{Model Comparison}
Two value-based methods are shown for comparison with the policy-based models described.
For both of these, the policy is implicitly determined by the action-value (Q) function which estimates the expected total return when choosing action $a$ given belief state $\mathbf{b}$ at time-step $t$. For an optimal policy $\pi^*$, the Q-function satisfies the {\it Bellman equation} \cite{bellman1954theory}:
\begin{equation}\label{eq:bellman}
\Scale[0.82]{Q^*(\mathbf{b}_t, a_t) = \text{E}_{\pi^*}  \{ r_t + \gamma \max_{a'} Q^*(\mathbf{b}_{t+1}, a') | \mathbf{b}_t, a_t\}.}
\end{equation}

\subsubsection{Deep Q-Network (DQN)}
DQN is a variant of the Q-learning algorithm whereby a neural network is used to non-linearly approximate the Q-function.
This suggests a sequential approximation
in Equation \ref{eq:bellman} by minimising the loss:
\begin{equation} \label{eq:dqnloss}
L(w_t) = \mathbb{E}\left[(y_t~ - Q(\mathbf{b}_t,a_t;w_t))^2 \right],
\end{equation}
where $y_t = r_t + \gamma \max_{a'}Q(\mathbf{b}_{t+1}, a';w^{-}_t)$ is the target to update the parameters $w$.
Note that $y_t$ is evaluated by a target network $w^{-}$ which is updated less frequently than the network $w$ to stabilise learning,
and the expectation is over the tuples $(\mathbf{b}_t, a_t, r_{t+1}, \mathbf{b}_{t+1})$ sampled from the experience replay pool described in \S \ref{sec:stabilising}. 

DQN often suffers from over-estimation on Q-values as the $\max$ operator is used to select an action as well as to evaluate it. 
Double DQN (DDQN) \cite{van2016deep} is thus used to de-couple the action selection and Q-value estimation to achieve better performance.
 
\subsubsection{Gaussian Processes (GP) RL}
GPRL is a state-of-the-art value-based RL algorithm for dialogue modelling.
It is appealing since it can learn from a small number of observations by exploiting the correlations defined by a {\it kernel function} and provides an uncertainty measure of its estimates. In GPRL, the $Q$-function is modelled as a GP with zero mean and kernel: $Q(B, A) \sim \mathcal{GP}(0, (k(\mathbf{b}, a), k(\mathbf{b}, a))$. This Q-function is then updated by calculating the posterior given the collected belief-action pairs $(\mathbf{b},a)$ (dictionary points) and their corresponding rewards \cite{GPRL}. The implicit knowledge of the distance between data points in  observation space provided by the kernel greatly speeds up  learning  since it enables Q-values in as yet unexplored space to be estimated.
Note that GPRL was used by ~\newcite{fatemi2016policy} to compare with deep RL but no uncertainty estimate was used to guide exploration and as a result had relatively poor performance. 
Here GPRL with uncertainty estimate is used as 
the benchmark.

\subsection{Reinforcement Learning from Scratch} \label{sec:RL}
The proposed models were first evaluated under 0\% semantic error rate with an agenda-based simulator which generates user interactions at the semantic-level~\cite{schatzmann2006survey}. In this case, the user intent is perfectly captured in the dialogue belief state without noise. 

The total return of each dialogue was set to $\mathds{1}(\mathcal{D})- 0.05 \times T$, where $T$ is the dialogue length and $\mathds{1}(\mathcal{D})$ is the success indicator for dialogue $\mathcal{D}$. The maximum dialogue length was set to 20 turns and $\gamma$ was 0.99. All deep RL models (A2C, TRACER, eNACER and DQN) contained two hidden layers of size 130 and 50.
The Adam optimiser was used \cite{kingma2014adam} with an initial learning rate of 0.001. During training, an $\epsilon$-greedy policy was used,
which was initially set to 0.3 and annealed to 0.0 over 3500 training dialogues. For GP, a linear kernel was used.

The ER pool ${\mathcal{P}}$ size was 1000, and the mini-batch size was 64. Once an initial 192 samples had been collected, the model was updated after every 2 dialogues. Note that for DQN, each sample was a state transition $(\mathbf{b}_t, a_t, r_t, \mathbf{b}_{t+1})$, whereas in A2C, TRACER and eNACER, each sample comprised the whole dialogue with all its state transitions. 
For eNACER, the natural gradient was computed to update the model weights of size $\sim$ 42000. 
For TRACER, $\alpha$ 
was set to 0.02, and $\xi$ was 0.01. 
Since the IS ratio 
has a high variance and can occasionally be extremely large, it was clipped between [0.8,1.0] to maintain stable training.

\label{sec:simulate}
\begin{figure}[t]
\centerline{\includegraphics[scale=0.42]{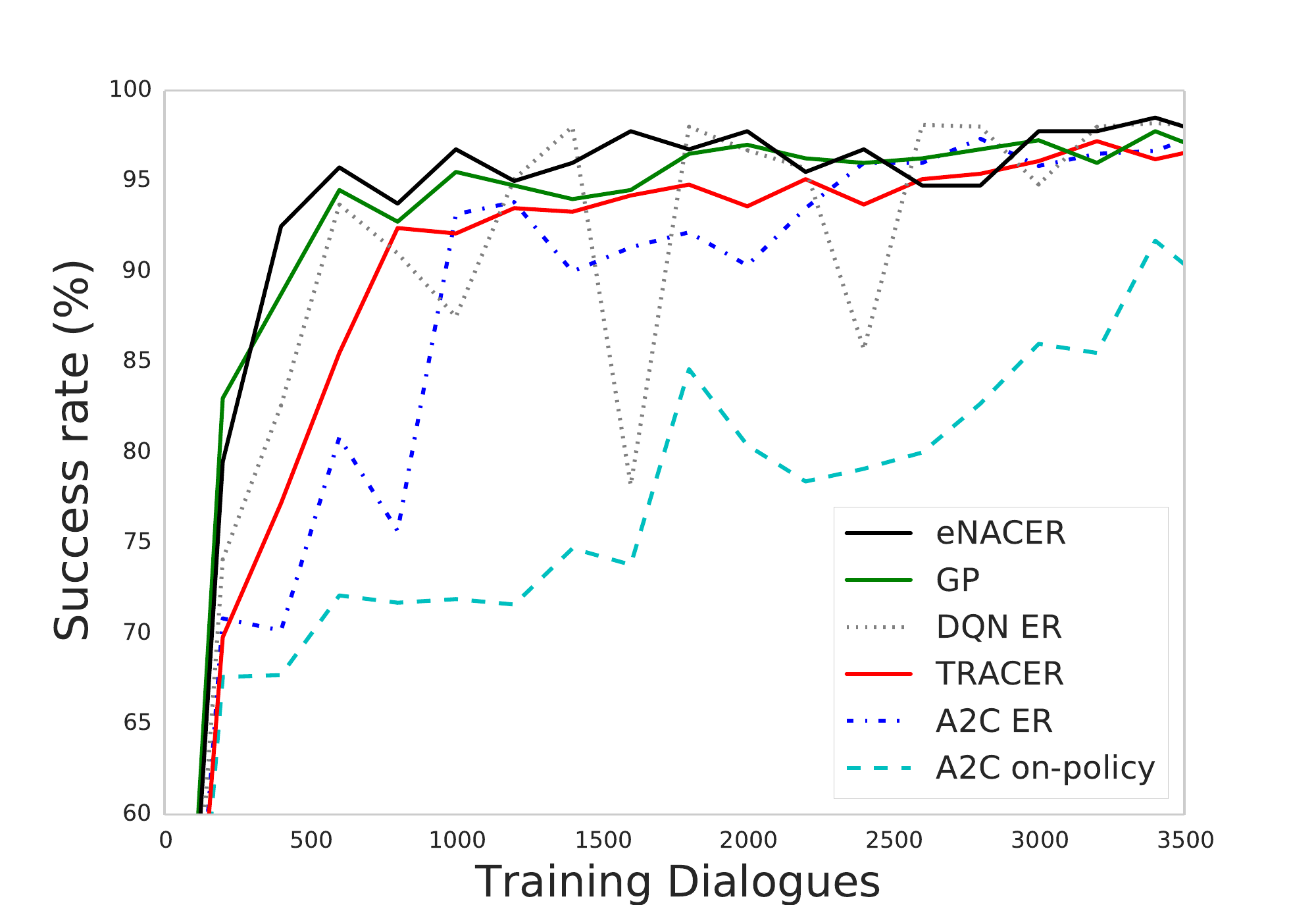}}
\caption{The success rate learning curves of on-policy A2C, A2C with ER, TRACER, DQN with ER, GP and eNACER in user simulation under noise-free condition.} 
\label{fig:DRL_curve}
\vspace{-2mm}
\end{figure}

Figure \ref{fig:DRL_curve} shows the success rate learning curves of on-policy A2C, A2C with ER, TRACER, DQN with ER, GP and eNACER. All were tested with 600 dialogues after every 200 training dialogues. 
As reported in previous studies, the benchmark GP model learns quickly and is relatively stable. eNACER provides comparable performance.  
DQN also showed high sample-efficiency but with high instability at some points. This
is because an iterative improvement in value space does not guarantee an improvement in policy space.
Although comparably slower to learn, the difference between on-policy A2C and A2C with ER clearly demonstrates the sample-efficiency of re-using past samples in mini-batches.
The enhancements incorporated into the TRACER algorithm do make this form of learning competitive although it still lags behind eNACER and GPRL.

\subsubsection{Learning from Demonstration Data} \label{sec:supervised}

\begin{figure*}[t!]
    \centering
    \begin{subfigure}[t]{0.5\textwidth}
        \centering
        \includegraphics[scale=0.4]{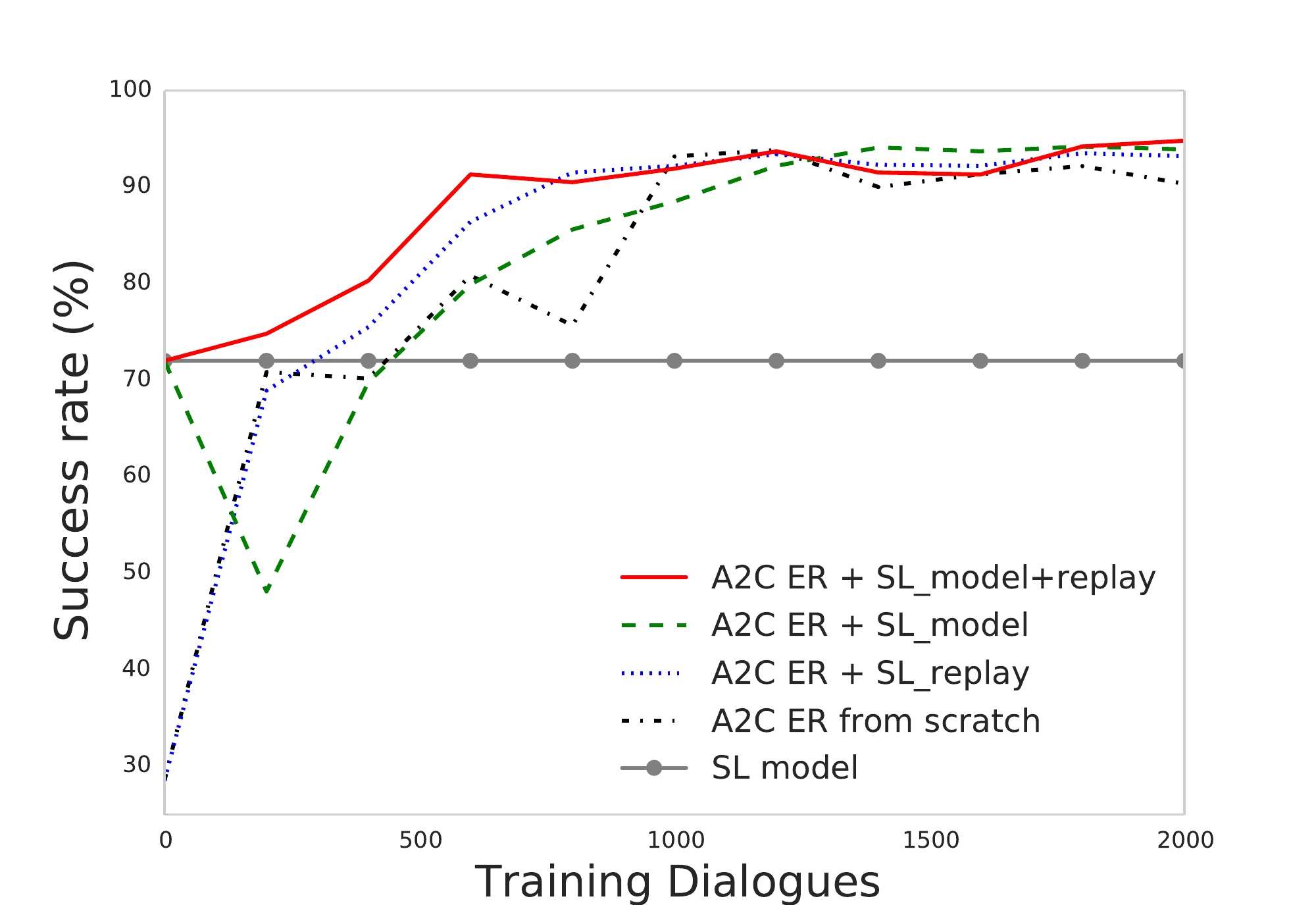}
        \caption{Learning for A2C ER with demonstration data.}
        \label{fig:a2c_slrl}
    \end{subfigure}%
    ~ 
    \begin{subfigure}[t]{0.5\textwidth}
        \centering
        \includegraphics[scale=0.4]{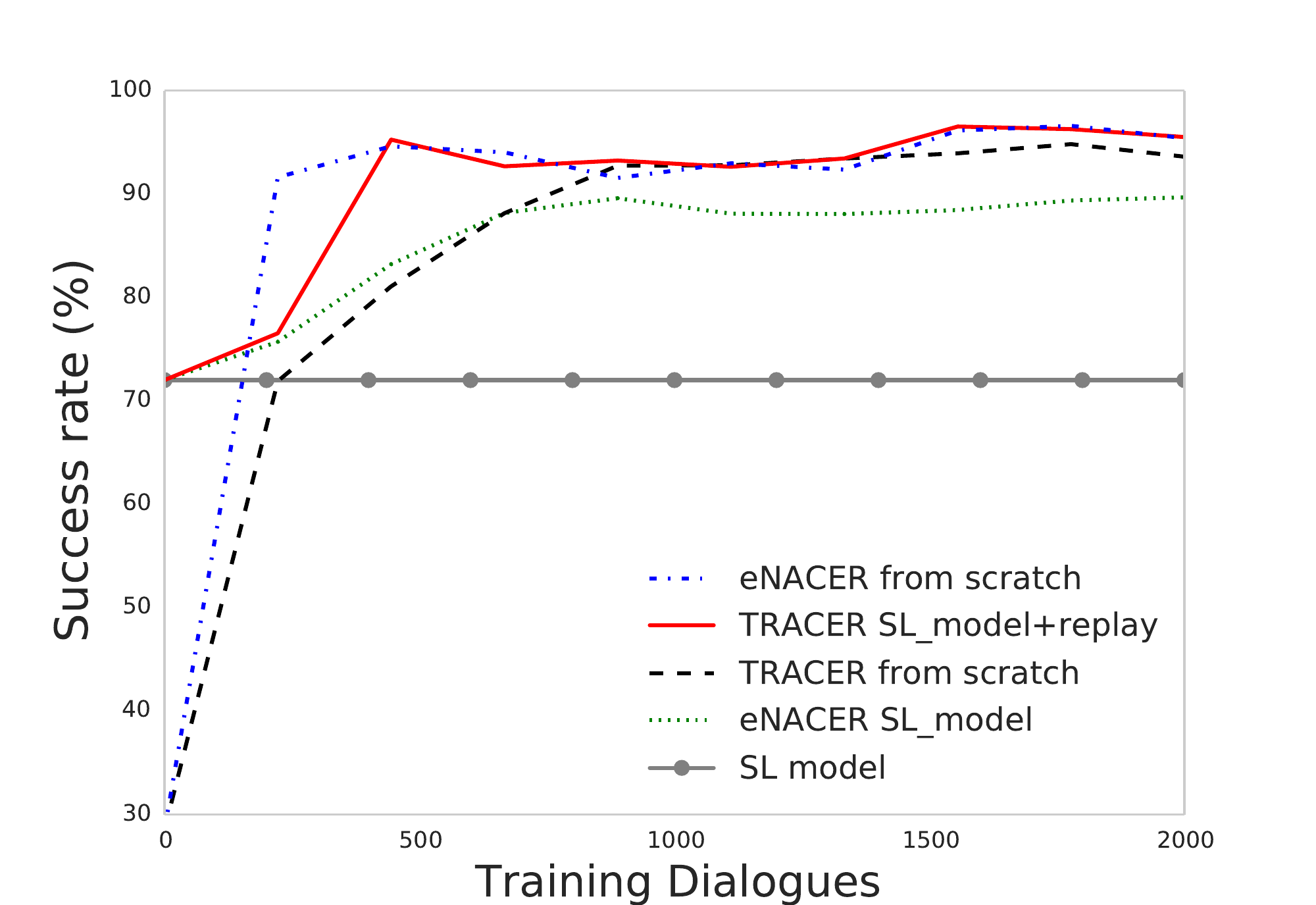}
        \caption{Learning for TRACER and eNACER with demonstration data.}
        \label{fig:a2c_tracer_slrl}
    \end{subfigure}
    \caption{Utilising demonstration data for improving RL learning speed.}
\end{figure*}

Regardless of the choice of  model and learning algorithm, training a policy from scratch on-line will always result in a poor user experience until sufficient interactions have been experienced to allow acceptable behaviours to be learned.

As discussed in \S \ref{sec:SLRL}, an off-line corpus of demonstration data can potentially mitigate this problem.  To test this, a corpus of 720 real user spoken dialogues in the Cambridge restaurant domain was utilised.
The corpus was split in a 4:1:1 ratio for training, validation and testing. It contains interactions between real users recruited via the Amazon Mechanical Turk service and a well-behaved SDS as described in \newcite{su:2016:acl}.

For A2C with ER and TRACER, the three ways of exploiting demonstration data in \S \ref{sec:SLRL} were explored.
The exploration parameter $\epsilon$ was also set to 0.3 and annealed to 0.0 over 2000 training dialogues. 
Since TRACER has similar patterns to A2C with ER, we first explored the impact of demonstration data on the A2C with ER results since it provides more headroom for identifying performance gains.

Figure \ref{fig:a2c_slrl} shows the different combinations of demonstration data using A2C with ER in noise-free conditions. The supervised pre-trained model (SL model) provides reasonable starting performance. 
The A2C ER model with supervised pre-training (A2C ER+SL\_model) improves on this after only 400 dialogues whilst suffering initially.
We hypothesise that the optimised SL pre-trained parameters distributed very differently to the optimal A2C ER parameters.
Also, the A2C ER model with SL replay (A2C ER+SL\_replay) shows clearly how the use of a supervised replay buffer can accelerate learning from scratch. Moreover, when SL pre-training is combined with SL replay
(A2C ER+SL\_model+replay), it achieved the best result.
Note that $\lambda_1$ and $\lambda_2$ in Equation \ref{eq:loss_all} were 10 and 0.01 respectively. In each policy update, 64 demonstration data were randomly sampled from the supervised replay pool ${\mathcal{P}_{sup}}$, which is the same number of RL samples selected from ER for A2C learning. 
Similar patterns emerge when utilising demonstration data to improve early learning in the TRACER and eNACER algorithms as shown in Figure \ref{fig:a2c_tracer_slrl}. However, in this case, eNACER is less able to exploit demonstration data 
since the training method is different from standard actor-critics. Hence, the  supervised loss $\mathcal{L}$ cannot be directly incorporated into the RL objective $J$ as in Equation \ref{eq:loss_all}. One could optimise the model using $\mathcal{L}$ separately after every RL update. However, in our experiments, this did not yield improvement. Hence, only eNACER learning from a pre-trained SL model is reported here. Compared to eNACER learning from scratch, eNACER from SL model started with good performance but
learned more slowly.
Again, this may be because the optimised SL pre-trained parameters distributed very differently from the optimal eNACER parameters and led to sub-optimality.
Overall, these results suggest that the proposed SL+RL framework to exploit demonstration data is effective in mitigating the cold start problem and TRACER provides the best solution in terms of avoiding poor initial performance, rapid learning and competitive fully trained performance.

\begin{figure}[t]
\centerline{\includegraphics[scale=0.32]{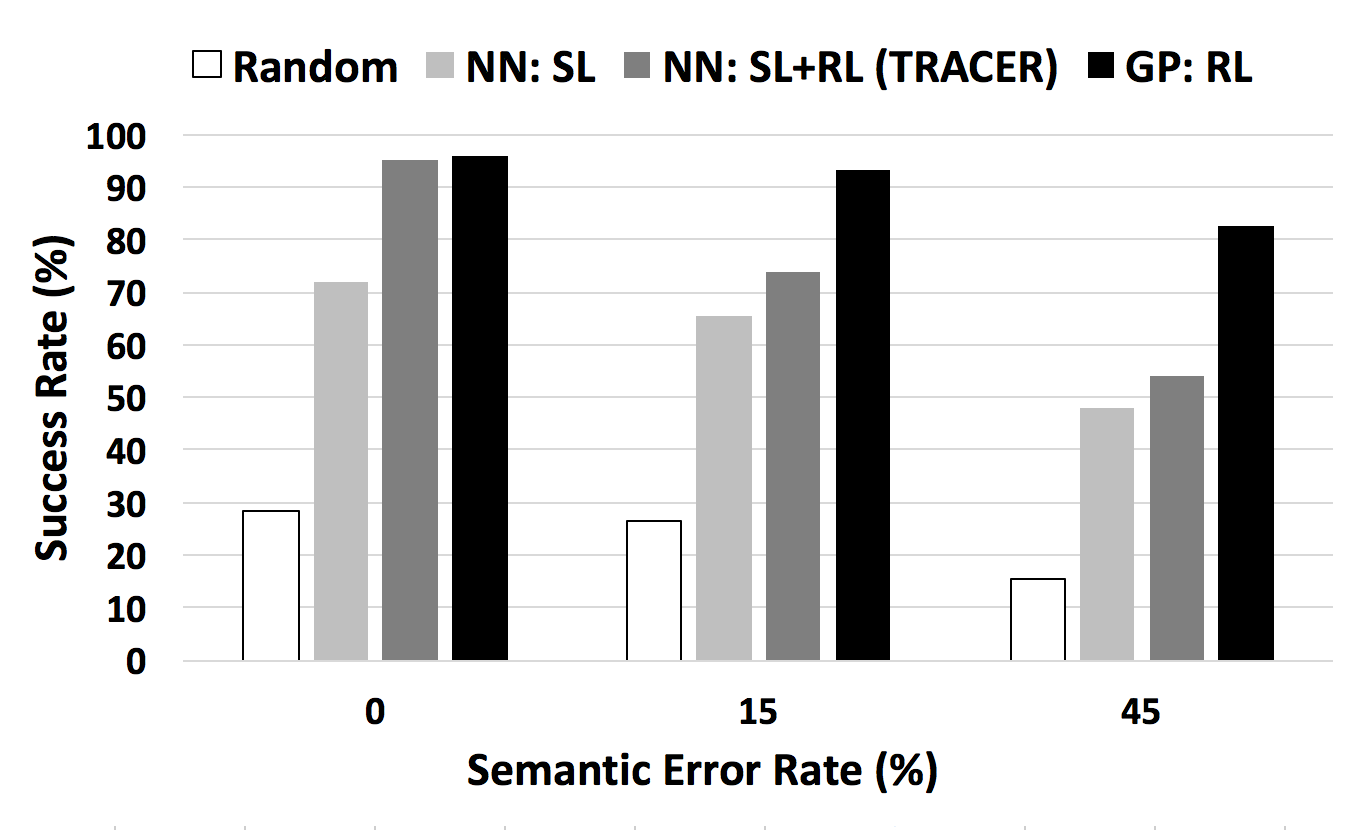}}
\caption{The success rate of TRACER for a random policy, policy trained with corpus data (NN:SL) and further improved via RL (NN:SL+RL) respectively in user simulation under various semantic error rates.} 
\label{fig:bar1}
\vspace{-5mm}
\end{figure}

In addition to the noise-free performance, we also investigated the impact of noise on the TRACER algorithm. 
Figure \ref{fig:bar1} shows the results after training on $2000$  dialogues via interaction with the user simulator under different semantic error rates.
The random policy (white bars) uniformly sampled an action from the set of size 14. 
This can be regarded as the average initial performance of any learning system.
We can see that SL generates a robust model which can be further fine-tuned using RL over a wide range of error rates. It should be noted, however, that the drop-off in performance at high noise levels is more rapid than might be expected, comparing to the GPRL. We believe that deep architectures are prone to overfitting and in consequence do not handle well the uncertainty of the user behaviour. We plan to investigate this issue in future work.
Overall, these outcomes validate the benefit of the proposed two-phased approach where the system can be effectively pre-trained using corpus data and further be refined via user interactions.

\section{Conclusion} \label{sec:conclude}
This paper has presented two compatible approaches to tackling the problem of slow learning and poor initial performance in deep reinforcement learning algorithms. Firstly, trust region actor-critic with experience replay (TRACER) and episodic natural actor-critic with experience replay (eNACER) were presented, these have been shown to be more sample-efficient than other deep RL models and broadly competitive with GPRL. Secondly, it has been shown that demonstration data can be utilised
to mitigate poor performance in the early stages of learning. To this end, 
two methods for using off-line corpus data were presented: simple pre-training using SL, and using the corpus data in a replay buffer.  These were particularly effective when used with TRACER which provided the best overall performance. 

Experimental results were also presented for mismatched environments, again TRACER demonstrated the ability to avoid poor initial performance when trained only on the demonstration corpus, yet still improve substantially with subsequent reinforcement learning. It was noted, however, that performance still falls off rather rapidly in noise compared to GPRL as the uncertainty estimates are not handled well by neural networks architectures.

Finally, it should be emphasised that whilst this paper has focused on the early stages of learning a new domain where GPRL provides a benchmark and is hard to beat, the potential of deep RL is its readily scalability to exploit on-line learning with large user populations as the model size is not related with experience replay buffer. 

\section*{Acknowledgments}
Pei-Hao Su is supported by Cambridge Trust and the Ministry of Education, Taiwan.
Pawe\l{} Budzianowski is supported by EPSRC Council and Toshiba Research Europe Ltd, Cambridge Research Laboratory.
The authors would like to thank the other members of the Cambridge Dialogue Systems Group for their valuable comments.

\bibliographystyle{acl2017}
\bibliography{acl2017}

\end{document}